\documentclass[11pt,a4paper,table]{article}
\usepackage[hyperref]{acl2019}
\usepackage[utf8]{inputenc}
\usepackage{times}
\usepackage{latexsym}
\usepackage{gnuplot-lua-tikz}
\usepackage{booktabs}
\usepackage{multirow}
\usepackage{array}

\usepackage{url}

\aclfinalcopy 
\def\aclpaperid{***} 

\title{CUNI System for the WMT19 Robustness Task}

\author{Jindřich Helcl \and Jindřich Libovický \and Martin Popel\\
  Charles University, Faculty of Mathematics and Physics, \\
  Institute of Formal and Applied Linguistics, \\
  Malostranské náměstí 25, 118 00 Prague, Czech Republic \\
  \texttt{\{helcl, libovicky, popel\}@ufal.mff.cuni.cz}
 \\}

\date{}

\begin{document}
\maketitle
\begin{abstract}
We present our submission to the WMT19 Robustness Task.
Our baseline system is the Charles University (CUNI) Transformer system
trained for the WMT18 shared task on News Translation.
Quantitative results show that the CUNI Transformer system is already far
more robust to noisy input than the LSTM-based baseline provided by the
task organizers.
We further improved the performance of our model by fine-tuning on the
in-domain noisy data without influencing the translation quality on the
news domain.
%
\end{abstract}

\section{Introduction}

Machine translation (MT) is usually evaluated on text coming from news
written by a professional journalist. However, in practice, MT should cover
more domains, including informal and not carefully spelled text that we
encounter in the online world.

Although the MT quality improved dramatically in recent
years \citep{bojar2018findings}, several studies
\citep{belinkov2018synthetic,khayrallah2018impact} has shown that the
current systems are sensitive to the source-side noise. It is also an issue
that was not studied intensively in the past because neural systems appear
to be more noise-sensitive than the previously used statistical systems
\citep{khayrallah2018impact}.

Recently, \citet{michel2018mtnt} prepared a dataset called Machine
Translation of Noisy Text (MTNT) that focuses exclusively on translating
texts from the online environment. This dataset is used for the WMT19
Robustness Task \citep{robustness:2019:WMT}.

\section{MTNT Dataset and Baselines}\label{sec:data}

The MTNT dataset consists of sentences collected from
Reddit\footnote{\url{http://www.reddit.com}} posts. Unlike the standard
corpora which (in a major part) consist of formal language, often written
by professionals, this dataset contains a substantial number of spelling
errors, grammatical errors, emoticons, and profanities.

Manual translations are provided along with the crawled source sentences.
The translators were asked to keep all the noise-related properties of
the source sentence.

There are two language pairs included in the dataset: English-French and
English-Japanese in both directions. The dataset comes in three splits, for
training, validation, and testing. The English-French part consists of 36k
examples in the training split, 852 examples for validation, 1020 examples for
testing in the En$\rightarrow$Fr direction, and 19k, 886, and 1022 examples
for training, validation, and testing respectively in the opposite direction.
For English-Japanese, the dataset is substantially smaller, with around 6k
training examples in both directions. In our experiments, we focus solely on
the translation between French and English.

We noticed that the MTNT dataset as provided for the task has some
peculiarities that were probably caused inadvertently during the dataset
building. Namely, the training and validation splits seem to come from a single
alphabetically sorted file. This means that all validation source sentences
start with the letter ``Y'', and anything that comes after ``Y'' in the
alphabetical order. Because of this, the validation scores are unreliable.
Moreover, a system trained on the training split will have a difficult time
translating sentences beginning with e.g. the word ``You'', which is a commonly
seen instance in the online discussion domain. This does not affect the test
split.

The baseline system introduced with the dataset is a recurrent
sequence-to-sequence model with attention \citep{bahdanau2014neural}. The
encoder is a bidirectional LSTM with two layers. The decoder is a two-layer
LSTM. The hidden state dimension in the LSTMs is 1,024 and the word embedding
size is 512.

The model that was used as a baseline for the Robustness Task was trained on
the WMT15 parallel data. Additionally, simple fine-tuning using stochastic
gradient descent on the MTNT data is shown to improve the translation quality
by a large margin. The translation quality of the system is tabulated among our
systems in Table~\ref{tab:results}.

\section{Related Work}

There have been several attempts to increase the robustness of MT systems in
recent years.

\citet{cheng2018towards} employ an adversarial training scheme in a multi-task
learning setup in order to increase the system robustness. For each training
example, its noisy counterpart is randomly generated. The network is trained to
yield such input representations such that it is not possible to train a
discriminator that decides (based on the input representation) which input is
the noisy one. This method improves both the robustness and the translation
quality on the clean data.

\citet{liu2018robust} attempt to make the translation more robust towards noise
from homophones. This type of noise is common in languages with non-phonetic
writing systems and concerns words or phrases which are pronounced in the same
way, but spelled differently. The authors of the paper train the word
embeddings to capture the phonetic information which eventually leads not only
to bigger robustness but also to improved translation quality in general.

To our knowledge, the only work that specifically uses the MTNT dataset
attempts to improve the system robustness by emulating the noise in the clean
data \citep{vaibhav2019improving}. They introduce two techniques for noise
induction, one employing hand-crafted rules, and one based on back-translation.
The techniques offer a similar translation quality gains as fine-tuning on MTNT
data.

\section{The CUNI Transformer model}

Our original plan was to train a system that would be robust by itself and
would not require further fine-tuning on the MTNT dataset.

Our baseline is the Transformer ``Big'' model
\citep{vaswani2017attention} as implemented in Tensor2Tensor
\citep{vaswani2018tensor2tensor}.
We train the model using the procedure
described in \citet{popel2018cuni} and \citet{popel2018training}, which was the
best-performing method for Czech-to-English and English-to-Czech translation in the
WMT18 News Translation shared task \citep{bojar2018findings}.

\begin{table}[t]

    \centering
    \begin{tabular}{@{}llr@{}}
    \toprule
    \multicolumn{2}{l}{Corpus} & \# Sentences \\ \midrule
    \multirow{5}{*}{\rotatebox{90}{Parallel}}
        & $10^9$ English-French Corpus & 22,520k \\
        & Europarl & 2,007k  \\
        & News Commentary & 200k \\
        & UN Corpus & 12,886k \\
        & Common Crawl & 3,224k \\ \midrule
    \multirow{2}{*}{\rotatebox{90}{Mono}}
        & French News Crawl ('08--'14) & 37,320k \\ 
        & English News Crawl ('11--'17) & 127,554k \\
        \bottomrule
    \end{tabular}

    \caption{Overview of the data used to train the CUNI Transformer baseline
    system.}\label{tab:data}

\end{table}

We trained our model on all parallel data available for the WMT15 News
Translation task \citep{bojar2015findings}.
We acquired additional synthetic
data by back-translation of the WMT News Crawl corpora
(from years 2008--2014 for French and 2011--2017 for English).
We did not include the News Discussion corpus that we considered too noisy for
training the system. Table~\ref{tab:data} gives an overview of the training
data composition.

\section{Fine-Tuning}

Similarly to the baseline experiments presented with the MTNT dataset
\citep{michel2018mtnt}, we fine-tune our general-domain model on the MTNT
dataset.

We continued the training of the models using the training part of the MTNT
dataset. Unlike the original model, we used plain stochastic gradient descent
with a constant learning rate for updating the weights. We executed several fine-tuning runs with different
learning rates and observed that learning rates smaller than $10^{-5}$ do not
change the model outputs at all and learning rates larger than $10^{-4}$ cause the models to diverge immediately.
The models in our final submission were fine-tuned with a learning rate of $10^{-4}$.

\begin{table*}[h]
    \centering
    \begin{tabular}{lcccccccc}
    \toprule
    & \multicolumn{4}{c}{English-French} & \multicolumn{4}{c}{French-English} \\
    \cmidrule(lr){2-5} \cmidrule(lr){6-9}
    & WMT14 & WMT15 & MTNT & blind
    & WMT14 & WMT15 & MTNT & blind \\
    \midrule
    MTNT baseline    & 33.5 & 33.0 & 21.8 & 22.1   & 28.9 & 30.8 & 23.3 & 25.6 \\
    + fine-tuning    & ---  & ---  & 29.7 & ---    & ---  & ---  & 30.3 & ---  \\
    \midrule
    CUNI Transformer & 43.6 & 41.6 & 34.0 & 37.0   & 42.9 & 39.6 & 39.9 & 42.6 \\
    + fine-tuning    & 43.5 & 41.6 & 36.6 & 38.5   & 41.5 & 40.9 & 42.1 & 44.8 \\
    \bottomrule
    \end{tabular}
    \caption{BLEU scores of the baseline and CUNI models measured on several
      datasets.}\label{tab:results}
\end{table*}

\begin{table}[t]
\centering

\begin{tabular}{lcc}\toprule
& en-fr & fr-en \\ \midrule
Naver Labs Europe               & 41.4 & 47.9 \\
\rowcolor{lightgray!25} this work                   & 38.5 & 44.8 \\
Baidu \& Oregon State Uni.      & 36.4 & 43.6 \\
Johns Hopkins Uni.	            &  --- & 40.2 \\
Fraunhofer FOKUS -- VISCOM	    & 24.2 & 29.9 \\
MTNT Baseline                   & 22.1 & 25.6 \\
\bottomrule
\end{tabular}

\caption{Quantiative comparison of the CUNI Transformer system + fine-tuning (this work) with other submitted systems.
}\label{tab:others}

\end{table}

\begin{figure}[t]

	\centering

    \input{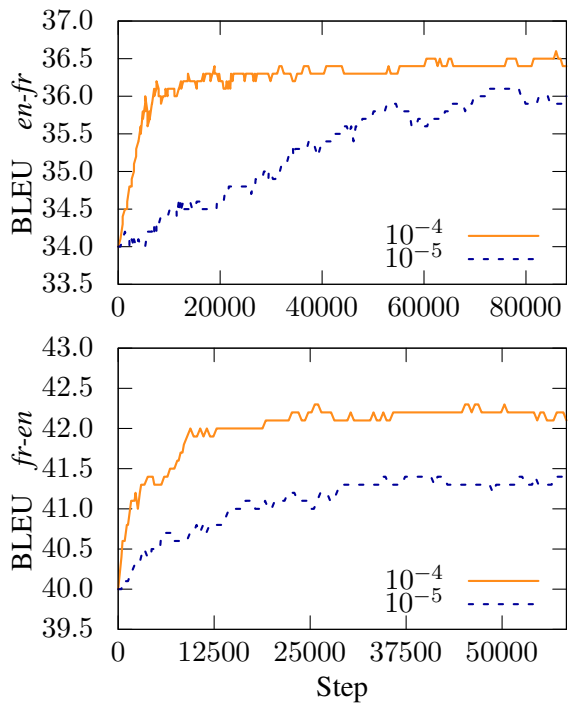}
    \begin{tikzpicture}[gnuplot]
\path (0.000,0.000) rectangle (7.000,4.500);
\gpcolor{color=gp lt color border}
\gpsetlinetype{gp lt border}
\gpsetdashtype{gp dt solid}
\gpsetlinewidth{1.00}
\draw[gp path] (0.920,0.770)--(1.100,0.770);
\draw[gp path] (6.815,0.770)--(6.635,0.770);
\node[gp node right] at (0.736,0.770) {$39.5$};
\draw[gp path] (0.920,1.303)--(1.100,1.303);
\draw[gp path] (6.815,1.303)--(6.635,1.303);
\node[gp node right] at (0.736,1.303) {$40.0$};
\draw[gp path] (0.920,1.835)--(1.100,1.835);
\draw[gp path] (6.815,1.835)--(6.635,1.835);
\node[gp node right] at (0.736,1.835) {$40.5$};
\draw[gp path] (0.920,2.368)--(1.100,2.368);
\draw[gp path] (6.815,2.368)--(6.635,2.368);
\node[gp node right] at (0.736,2.368) {$41.0$};
\draw[gp path] (0.920,2.901)--(1.100,2.901);
\draw[gp path] (6.815,2.901)--(6.635,2.901);
\node[gp node right] at (0.736,2.901) {$41.5$};
\draw[gp path] (0.920,3.434)--(1.100,3.434);
\draw[gp path] (6.815,3.434)--(6.635,3.434);
\node[gp node right] at (0.736,3.434) {$42.0$};
\draw[gp path] (0.920,3.966)--(1.100,3.966);
\draw[gp path] (6.815,3.966)--(6.635,3.966);
\node[gp node right] at (0.736,3.966) {$42.5$};
\draw[gp path] (0.920,4.499)--(1.100,4.499);
\draw[gp path] (6.815,4.499)--(6.635,4.499);
\node[gp node right] at (0.736,4.499) {$43.0$};
\draw[gp path] (0.920,0.770)--(0.920,0.950);
\draw[gp path] (0.920,4.499)--(0.920,4.319);
\node[gp node center] at (0.920,0.462) {$0$};
\draw[gp path] (2.182,0.770)--(2.182,0.950);
\draw[gp path] (2.182,4.499)--(2.182,4.319);
\node[gp node center] at (2.182,0.462) {$12500$};
\draw[gp path] (3.445,0.770)--(3.445,0.950);
\draw[gp path] (3.445,4.499)--(3.445,4.319);
\node[gp node center] at (3.445,0.462) {$25000$};
\draw[gp path] (4.707,0.770)--(4.707,0.950);
\draw[gp path] (4.707,4.499)--(4.707,4.319);
\node[gp node center] at (4.707,0.462) {$37500$};
\draw[gp path] (5.969,0.770)--(5.969,0.950);
\draw[gp path] (5.969,4.499)--(5.969,4.319);
\node[gp node center] at (5.969,0.462) {$50000$};
\draw[gp path] (0.920,4.499)--(0.920,0.770)--(6.815,0.770)--(6.815,4.499)--cycle;
\node[gp node center,rotate=-270] at (-0.308,2.634) {BLEU\quad \it fr-en};
\node[gp node center] at (3.867,0.000) {Step};
\node[gp node right] at (5.347,1.412) {$10^{-4}$};
\gpcolor{rgb color={1.000,0.549,0.102}}
\gpsetlinewidth{2.00}
\draw[gp path] (5.531,1.412)--(6.447,1.412);
\draw[gp path] (0.920,1.303)--(0.963,1.729)--(0.977,1.942)--(1.006,1.942)--(1.034,2.155)%
  --(1.049,2.155)--(1.091,2.475)--(1.093,2.475)--(1.136,2.475)--(1.148,2.581)--(1.179,2.368)%
  --(1.205,2.581)--(1.222,2.688)--(1.261,2.688)--(1.266,2.688)--(1.309,2.794)--(1.318,2.794)%
  --(1.352,2.794)--(1.375,2.794)--(1.395,2.688)--(1.431,2.688)--(1.438,2.688)--(1.482,2.688)%
  --(1.488,2.688)--(1.525,2.794)--(1.545,2.794)--(1.568,2.794)--(1.602,2.901)--(1.611,2.901)%
  --(1.654,2.901)--(1.659,2.901)--(1.697,3.007)--(1.715,3.007)--(1.741,3.114)--(1.766,3.114)%
  --(1.784,3.220)--(1.827,3.327)--(1.870,3.434)--(1.913,3.327)--(1.956,3.327)--(1.999,3.434)%
  --(2.042,3.327)--(2.086,3.434)--(2.129,3.327)--(2.172,3.327)--(2.215,3.434)--(2.258,3.434)%
  --(2.301,3.434)--(2.344,3.434)--(2.387,3.434)--(2.430,3.434)--(2.473,3.434)--(2.516,3.434)%
  --(2.559,3.434)--(2.602,3.434)--(2.646,3.434)--(2.689,3.434)--(2.732,3.434)--(2.775,3.434)%
  --(2.818,3.434)--(2.861,3.540)--(2.904,3.540)--(2.947,3.540)--(2.990,3.540)--(3.033,3.540)%
  --(3.076,3.540)--(3.119,3.540)--(3.162,3.540)--(3.205,3.647)--(3.248,3.647)--(3.291,3.647)%
  --(3.334,3.540)--(3.377,3.540)--(3.420,3.647)--(3.463,3.647)--(3.506,3.753)--(3.549,3.753)%
  --(3.592,3.647)--(3.635,3.647)--(3.677,3.647)--(3.720,3.647)--(3.763,3.540)--(3.806,3.540)%
  --(3.849,3.540)--(3.892,3.540)--(3.935,3.540)--(3.978,3.647)--(4.021,3.540)--(4.064,3.540)%
  --(4.107,3.540)--(4.150,3.540)--(4.193,3.540)--(4.235,3.540)--(4.278,3.647)--(4.321,3.540)%
  --(4.364,3.540)--(4.407,3.647)--(4.450,3.540)--(4.493,3.540)--(4.537,3.647)--(4.580,3.647)%
  --(4.623,3.647)--(4.666,3.647)--(4.709,3.647)--(4.752,3.647)--(4.795,3.647)--(4.838,3.647)%
  --(4.881,3.647)--(4.924,3.647)--(4.967,3.647)--(5.010,3.647)--(5.053,3.647)--(5.096,3.647)%
  --(5.139,3.647)--(5.182,3.647)--(5.225,3.647)--(5.268,3.647)--(5.311,3.647)--(5.354,3.647)%
  --(5.397,3.647)--(5.440,3.647)--(5.483,3.753)--(5.526,3.753)--(5.569,3.647)--(5.612,3.753)%
  --(5.656,3.753)--(5.699,3.647)--(5.742,3.647)--(5.785,3.647)--(5.828,3.647)--(5.871,3.647)%
  --(5.914,3.647)--(5.957,3.647)--(6.000,3.753)--(6.043,3.647)--(6.086,3.647)--(6.129,3.647)%
  --(6.172,3.647)--(6.215,3.647)--(6.259,3.647)--(6.302,3.647)--(6.345,3.647)--(6.388,3.647)%
  --(6.431,3.540)--(6.473,3.647)--(6.516,3.647)--(6.559,3.540)--(6.602,3.540)--(6.645,3.540)%
  --(6.688,3.647)--(6.731,3.647)--(6.775,3.647)--(6.815,3.540);
\gpcolor{color=gp lt color border}
\node[gp node right] at (5.347,1.104) {$10^{-5}$};
\gpcolor{rgb color={0.000,0.000,0.600}}
\gpsetdashtype{dash pattern=on 2.00*\gpdashlength off 5.00*\gpdashlength }
\draw[gp path] (5.531,1.104)--(6.447,1.104);
\draw[gp path] (0.920,1.303)--(0.963,1.303)--(0.975,1.409)--(1.006,1.409)--(1.030,1.409)%
  --(1.049,1.409)--(1.085,1.516)--(1.092,1.516)--(1.135,1.622)--(1.140,1.622)--(1.178,1.622)%
  --(1.195,1.622)--(1.221,1.729)--(1.250,1.835)--(1.264,1.835)--(1.305,1.729)--(1.307,1.729)%
  --(1.350,1.729)--(1.360,1.835)--(1.393,1.835)--(1.415,1.835)--(1.436,1.835)--(1.470,1.835)%
  --(1.479,1.942)--(1.522,2.049)--(1.525,2.049)--(1.565,2.049)--(1.580,2.049)--(1.608,2.049)%
  --(1.635,1.942)--(1.651,1.942)--(1.690,1.942)--(1.693,1.942)--(1.736,1.942)--(1.745,1.942)%
  --(1.766,1.942)--(1.779,1.942)--(1.822,1.942)--(1.865,2.049)--(1.908,2.049)--(1.951,2.155)%
  --(1.994,2.049)--(2.037,2.155)--(2.079,2.049)--(2.122,2.155)--(2.165,2.155)--(2.208,2.155)%
  --(2.251,2.155)--(2.294,2.155)--(2.337,2.262)--(2.380,2.368)--(2.423,2.368)--(2.466,2.368)%
  --(2.509,2.368)--(2.551,2.368)--(2.594,2.475)--(2.637,2.475)--(2.680,2.475)--(2.723,2.368)%
  --(2.766,2.368)--(2.809,2.368)--(2.852,2.475)--(2.894,2.368)--(2.937,2.368)--(2.980,2.475)%
  --(3.023,2.475)--(3.066,2.475)--(3.109,2.475)--(3.151,2.581)--(3.194,2.581)--(3.237,2.581)%
  --(3.280,2.475)--(3.323,2.475)--(3.366,2.475)--(3.408,2.475)--(3.451,2.368)--(3.494,2.368)%
  --(3.537,2.475)--(3.580,2.581)--(3.623,2.581)--(3.666,2.581)--(3.708,2.475)--(3.751,2.475)%
  --(3.794,2.581)--(3.837,2.688)--(3.880,2.688)--(3.922,2.688)--(3.965,2.688)--(4.008,2.688)%
  --(4.051,2.688)--(4.094,2.688)--(4.136,2.688)--(4.179,2.688)--(4.222,2.688)--(4.265,2.688)%
  --(4.307,2.688)--(4.350,2.688)--(4.393,2.794)--(4.436,2.794)--(4.478,2.688)--(4.521,2.688)%
  --(4.564,2.688)--(4.607,2.688)--(4.650,2.688)--(4.693,2.794)--(4.735,2.794)--(4.778,2.794)%
  --(4.821,2.794)--(4.864,2.794)--(4.906,2.794)--(4.949,2.794)--(4.992,2.794)--(5.035,2.794)%
  --(5.078,2.688)--(5.120,2.794)--(5.163,2.794)--(5.206,2.794)--(5.249,2.688)--(5.291,2.688)%
  --(5.334,2.688)--(5.377,2.688)--(5.420,2.688)--(5.463,2.688)--(5.505,2.688)--(5.548,2.688)%
  --(5.591,2.688)--(5.634,2.688)--(5.677,2.688)--(5.720,2.688)--(5.763,2.688)--(5.805,2.688)%
  --(5.848,2.581)--(5.891,2.688)--(5.934,2.688)--(5.977,2.688)--(6.020,2.688)--(6.063,2.688)%
  --(6.106,2.688)--(6.149,2.688)--(6.191,2.794)--(6.234,2.794)--(6.277,2.688)--(6.320,2.688)%
  --(6.363,2.688)--(6.405,2.794)--(6.448,2.688)--(6.491,2.688)--(6.534,2.688)--(6.577,2.688)%
  --(6.620,2.688)--(6.662,2.794)--(6.705,2.794)--(6.748,2.794)--(6.791,2.794)--(6.815,2.794);
\gpcolor{color=gp lt color border}
\gpsetdashtype{gp dt solid}
\gpsetlinewidth{1.00}
\draw[gp path] (0.920,4.499)--(0.920,0.770)--(6.815,0.770)--(6.815,4.499)--cycle;
\gpdefrectangularnode{gp plot 1}{\pgfpoint{0.920cm}{0.770cm}}{\pgfpoint{6.815cm}{4.499cm}}
\end{tikzpicture}

    \caption{Learning curves showing the progress of fine-tuning on the MTNT
    test split for English-to-French (top) and French-to-English (bottom) systems with two different learning rates.}
    %

\label{fig:curves}
\end{figure}


\section{Results}

We evaluate the results on four datasets. The first one is \emph{neswtest2014}
\citep{bojar2014findings}, a standard WMT test set consisting of manually
translated newspaper texts where one half is originally in English and the
other half originally in French.

Because of the large amount of training data available, even the statistical MT
systems achieved high translation quality on the news domain. Because of that, a
slightly different test set (\emph{newsdiscusstest2015}) was used as
the evaluation test set for the WMT15 competition \citep{bojar2015findings}.
The test set consists of sentences from discussions under news stories from The
Guardian and Le Monde. Even though the topics are the same as the news stories,
the language used in the discussions is less formal and contains grammatical
and spelling errors, which makes them somewhat closer to the MTNT dataset.

Finally, we evaluate the models on the test part of the MTNT dataset (described
in Section~\ref{sec:data}) and the blind test set for the WMT19 Robustness Task,
which was collected in the same way as the original MTNT dataset.

The quantitative results are shown in Table~\ref{tab:results}. The
Transformer-based baseline outperforms the RNN-based MTNT baseline by a large
margin on both WMT and MTNT test datasets.

The fine-tuning of the RNN-based models brings a substantial translation
quality boost of 8 and 7 BLEU points in each direction respectively. This
effect is much smaller with our stronger baseline and only improves the
performance by around 2 BLEU points in either direction. This may indicate that
sufficiently strong models are robust enough and do not need further
fine-tuning for the type of noise present in the MTNT dataset. Especially in
French-to-English translation, the fine-tuning improvement is reached at the
expense of decreased translation quality in the news domain.

We
observe that the fine-tuning has only a small negative impact on the
translation quality of our models on the general-domain data. It would be
interesting to see how big impact made the fine-tuning of the MTNT baseline
model, which gained such a large improvement on the domain-specific
data. However, the authors of the baseline \citep{michel2018mtnt}
do not report these results.

We plot the learning curves from the progress of the system fine-tuning in
Figure~\ref{fig:curves}. Even though the fine-tuning improved the model
performance on both language pairs by approximately the same margin, the
courses of the fine-tuning differ fundamentally. For English-to-French
translation, we see that the translation quality slowly increases until
convergence. For the opposite direction, it improves immediately and keeps
oscillating during the remaining training steps. We found that this effect was
similar regardless of the learning rate.

Although we observed a strong effect of checkpoint averaging during the
baseline model training, it has almost no effect on the fine-tuned
models. Therefore, we report only the performance for parameter checkpoints
with the highest validation BLEU scores.


Table~\ref{tab:others} compares the automatic scores with other WMT19
Robustness Task participants. Our submission was
outperformed by submissions by Naver Labs Europe in both translation
directions. Their submission used the same architecture as our submission, but
in addition, it employed corpus tags and synthetic noise generation. Details
about other systems were not known at the time of our submission.

\section{Conclusions}

In our submission to the WMT19 Robustness Task, we experiment with fine-tuning
of strong Transformer-based baselines for translation between English and
French.

Our results show that when using a strong baseline, the effect of fine-tuning
on a domain-specific dataset is much smaller than for weaker models introduced
as a baseline with the MTNT dataset.

\section*{Acknowledgements}

This research has been supported by the from the European Union's Horizon 2020 research and innovation programme under grant agreement No.~825303 (Bergamot),
Czech Science Foundtion grant No.~19-26934X (NEUREM3), and Charles University grant No.~976518,
and has been using language resources distributed by the LINDAT/CLARIN project of the Ministry of Education, Youth and Sports of the Czech Republic (LM2015071).

\bibliography{main}
\bibliographystyle{acl_natbib}

\end{document}